\documentclass{article} 
\usepackage{iclr2019_conference,times}

\usepackage{amsmath,amsfonts,bm}

\def\eqref#1{equation~\ref{#1}}

\def\1{\bm{1}}

\DeclareMathAlphabet{\mathsfit}{\encodingdefault}{\sfdefault}{m}{sl}
\SetMathAlphabet{\mathsfit}{bold}{\encodingdefault}{\sfdefault}{bx}{n}

\usepackage[utf8]{inputenc} 
\usepackage[T1]{fontenc}    
\usepackage{hyperref}       
\usepackage{url}            
\usepackage{booktabs}       
\usepackage{nicefrac}       
\usepackage{microtype}      
\usepackage{pifont}

\usepackage{wrapfig}
\usepackage{changepage}
\usepackage{amsmath}
\usepackage{amsthm}
\usepackage{amssymb}
\usepackage{amsfonts,eucal,amsbsy}
\usepackage{mathtools}
\usepackage{multirow}
\usepackage[normalem]{ulem}

\usepackage{subfigure}

\newcommand{\ignore}[1]{}
\newcommand{\vect}[1]{\mathbf{#1}}

\title{Variational Smoothing in Recurrent Neural Network Language Models}

\iclrfinalcopy
\author{Lingpeng Kong, Gabor Melis, Wang Ling, Lei Yu, Dani Yogatama\\
DeepMind\\
\texttt{\{lingpenk, melisgl, lingwang, leiyu, dyogatama\}@google.com} \\
}

\begin{document}

\maketitle

\begin{abstract}
We present a new theoretical perspective of data noising in
recurrent neural network language models \citep{xie}.
We show that each variant of data noising is an 
instance of Bayesian recurrent neural
networks with a particular variational distribution (i.e., 
a mixture of Gaussians whose weights depend on statistics derived from the corpus such as the unigram distribution).
We use this insight to propose a more principled 
method to apply at prediction time
and propose natural extensions to data noising under the variational framework.
In particular, we propose variational smoothing 
with tied input and output embedding matrices
and an element-wise variational smoothing method.
We empirically verify our analysis on two benchmark 
language modeling datasets and demonstrate 
performance improvements over existing data noising methods.
\end{abstract}

\section{Introduction}
Recurrent neural networks (RNNs) are state of the art models in various 
language processing tasks. However, 
their performance heavily depends on proper regularization 
at training time \citep{melisgl, smerity}.
The two predominant approaches to regularize RNNs are dropout (randomly zeroing out neurons; Srivastava et al., 2014) \nocite{dropout} and $\ell_2$ regularization (applying $\ell_2$ penalty to model parameters; Hoerl \& Kennard, 1970).  \nocite{hoerl}
Recently, \citet{xie} proposed data noising to regularize language models. 
Their method is formulated as a data augmentation method
that randomly replaces words with other words drawn from a proposal distribution. 
For example, we can use the unigram distribution which 
models the number of 
word occurrences in the training corpus, or a more sophisticated
proposal distribution that takes into 
account the number of bigram types in the corpus.
Data noising has been shown to improve perplexity 
on small and large corpora, and that this improvement is complementary to other regularization techniques and translates to improvements on downstream models
such as machine translation.

\citet{xie} derived connections between data noising and smoothing
in classical $n$-gram language models, which we review in \S{\ref{sec:datanoising}}.
In smoothing \citep{chen}, since empirical counts for unseen sequences are zero,
we smooth our estimates by a weighted average of higher order
and lower order $n$-gram models.
There are various ways to choose the weights
and the lower order models
leading to different smoothing techniques, with Kneser-Ney smoothing
widely considered to be the most effective.
\citet{xie} showed that the pseudocounts of a noised data correspond
to a mixture of different $n$-gram models.

In this paper, we provide a new theoretical foundation for data noising
and show that it can be understood as a form of Bayesian recurrent neural
network with a particular variational distribution (\S{\ref{sec:linint}} and \S{\ref{sec:knsmooth}}).
Our derivation relates data noising to dropout 
and variational dropout \citep{vardropout},
and naturally leads to a data dependent $\ell_2$ regularization coefficient.
We use this insight to arrive at a more principled way to do prediction 
with data noising---i.e., by taking the mean of the variational distribution,
as opposed to the mode---and propose several 
extensions under the variational framework in \S{\ref{sec:variants}}. Specifically, 
we show how to use variational smoothing with tied 
input and output embeddings and propose element-wise smooothing.
In \S{\ref{sec:exp}}, we validate our analysis in language modeling experiments 
on the Penn Treebank \citep{marcus1994penn} and Wikitext-2 \citep{merity2016pointer} datasets.

\section{Recurrent Neural Network Language Models}
We consider a language modeling problem where the goal is to predict the next word $x_t$ 
given previously seen context words $x_{<t} = \{x_0, x_1, \ldots, x_{t-1}\}$.
Let $\vect{W}$ be parameters of a recurrent neural network, and 
$p(x_t \mid x_{<t}; \vect{W}) = \text{RNN}(x_{<t}; \vect{W})$.
Following previous work in language modeling \citep{melisgl,smerity}, we use LSTM \citep{lstm} as our RNN function,
although other variants such as GRU \citep{gru} can be used as well.

Given a training corpus $\mathcal{D} = \{x_0, \ldots, x_T\}$, the likelihood we would like to maximize is:
\begin{align*}
\mathcal{L}(\vect{W}; \mathcal{D}) = \prod_{t=1}^T p(x_t \mid x_{<t}; \vect{W}).
\end{align*}

Directly optimizing the (log) likelihood above often leads to overfitting.
We typically augment the objective function with a regularizer 
(e.g., $\ell_2$ regularizer where $\mathcal{R}(\vect{W}) = \lambda \Vert \vect{W} \Vert_2^2$)
or use dropout by randomly zeroing out neurons \citep{dropout}.

\paragraph{Data noising as smoothing.}
\label{sec:datanoising}
\citet{xie} proposed a method to regularize recurrent 
neural network language models 
by noising the data.
For each input word $i$ in $x_{<t}$ (sometimes also the corresponding output word),
we replace it
with another word sampled from a proposal distribution $\mathcal{T}$
with probability $\gamma_i$.
They introduced various methods on how to choose $\gamma_i$ and $\mathcal{T}$.

For example, if $\gamma_i = \gamma$ for all $i \in \mathcal{V}$ 
and $\mathcal{T}$ is the unigram distribution of words in the corpus, 
$\mathbb{E}_{\mathcal{T}}[p(x_t \mid \text{noise}_{\gamma,\mathcal{T}}(x_{<t}); \vect{W})]$ corresponds to a mixture of $n$-gram models 
with fixed weights (i.e., linear interpolation smoothing).

Another option is to set $\gamma_i = \gamma \frac{\text{distinct}(i, \bullet)}{\text{count}(i)}$, where $\text{distinct}(i, \bullet)$ denotes the number of distinct continuations preceded by word $i$ (i.e., the number of bigram types that has $i$ as the first word), and $\text{count}(i)$ denotes the number of times $i$ appears in the corpus.
For this choice of $\gamma_i$, when $\mathcal{T}$ is the unigram distribution, the expectation corresponds to absolute discounting; whereas if $\mathcal{T}_i = \frac{\text{distinct}(\bullet, i)}{\sum_{v \in \mathcal{V}} \text{distinct}(\bullet, v)}$ 
and we replace both the input and output words, 
it corresponds to bigram Kneser-Ney smoothing.
We summarize their proposed methods in Table~\ref{tbl:smoothing}.

At prediction (test) time, \citet{xie} do not apply any noising and directly predict $p(x_t \mid x_{<t}; \vect{W})$.
They showed that a combination of smoothing and dropout achieves the best result on their language modeling and machine translation experiments.

\renewcommand{\arraystretch}{1.4}
\begin{table}[h]
\centering
\begin{tabular}{|c|r|r|r|}
\hline
Name & Noised & $\gamma_i$ & $\mathcal{T}_x$ \\ 
\hline
\hline
Blank noising & $x_0$ & $\gamma$ & $\mathcal{T}_{\text{``\_''}} = 1$\\
Linear interpolation & $x_0$ & $\gamma$ & unigram\\
Absolute discounting & $x_0$ & $\gamma \frac{\text{distinct}(i, \bullet)}{\text{count}(i)}$ & unigram\\
Kneser-Ney & $x_0, x_1$ & $\gamma \frac{\text{distinct}(i, \bullet)}{\text{count}(i)}$ & $\frac{\text{distinct}(\bullet, i)}{\sum_{v \in \mathcal{V}} \text{distinct}(\bullet, v)}$\\
\hline
\end{tabular}
\caption{Variants of data noising techniques proposed in \citet{xie} for context word $x_0$ and target word $x_1$. Blank noising replaces an input word with a blank word, denoted by ``\_''.
\label{tbl:smoothing}}
\end{table}

\section{Bayesian Recurrent Neural Networks}
\label{sec:brnn}
In Bayesian RNNs, we define a prior over our parameters $\vect{W}$ and consider the following maximization problem:
\begin{align*}
\mathcal{L}(\mathcal{D}) = \int \prod_{t=1}^T p(x_t \mid x_{<t}; \vect{W})p(\vect{W}) d_{\vect{W}}.
\end{align*}
A common prior is the standard normal distribution $p(\vect{W}) = \mathcal{N}(\vect{0}, \vect{I})$.

For recurrent neural networks, the posterior over $\vect{W}$ given the data $\mathcal{D}$ is intractable.
We approximate the posterior with a variational distribution $q(\vect{W})$, 
and minimize the KL divergence between the variational distribution and the true posterior by:
\begin{align*}
    \mathbb{KL} (q(\vect{W}) \Vert p(\vect{W}\mid\mathcal{D}) ) &= \mathbb{KL}(q(\vect{W}) \Vert p(\vect{W}) ) - \int \log p(\mathcal{D} \mid \vect{W}) q(\vect{W})d_{\vect{W}} + \text{constant}\\
    &= \mathbb{KL}(q(\vect{W}) \Vert p(\vect{W})) - \sum_{t=1}^T \int \log p(x_t \mid x_{<t}; \vect{W}) q(\vect{W})d_{\vect{W}} + \text{constant}
\end{align*}
The integral is often approximated with Monte Carlo integration with one sample $\vect{\tilde{W}} \sim q(\vect{W})$:\footnote{In the followings, we use $\tilde{x}$ to denote a sample from a distribution $q(x)$.}
\begin{align}
\label{eq:sampling}
\int \log p(x_t \mid x_{<t}; \vect{W}) q(\vect{W})d_{\vect{W}} \approx \log p(x_t \mid x_{<t}; \vect{\tilde{W}}) q(\vect{\tilde{W}})d_{\vect{\tilde{W}}}.
\end{align}

At test time, for a new sequence $y_0, \ldots, y_T$, we can either set $\vect{W}$ to be the mean of $q(\vect{W})$ or sample and average the results: $p(y_t \mid y_{<t}) = \frac{1}{S} \sum_{s=1}^S p(y_t \mid y_{<t}; \vect{\tilde{W}}_s)$, where $S$ is 
the number of samples and $\tilde{\vect{W}}_s \sim q(\vect{W})$.

\section{Variational Smoothing}
We now provide theoretical justifications for data smoothing under the variational framework.
In a recurrent neural network language model, there are three types of parameters: an input (word) embedding matrix $\vect{E}$, an LSTM parameter matrix $\vect{R}$, and an output embedding matrix $\vect{O}$ that produces logits for the final softmax function.
We have $\vect{W} = \{\vect{E}, \vect{R}, \vect{O}\}$.

\subsection{Linear interpolation smoothing.}
\label{sec:linint}
We first focus on the simplest data noising method---linear interpolation smoothing---and show how to extend it to other data noising methods subsequently.
\paragraph{Word embedding matrix $\vect{E}$.}
Denote the word vector corresponding to word $i \in \mathcal{V}$ in 
the input embedding matrix $\vect{E}$ by $\vect{e}_i$.
We obtain a similar effect to linear interpolation smoothing by using the 
following mixture of Gaussians variational distribution for $\vect{e}_i$:
\begin{align}
\label{eq:vardist}
\vect{e}_i \sim q(\vect{e}_i) &= (1 - \gamma) \mathcal{N}(\vect{e}_i, \sigma\vect{I})  + \gamma \sum_{v \in \mathcal{V}} \mathcal{U}_v \mathcal{N}(\vect{e}_v, \sigma\vect{I}) \nonumber \\
&= (1 - \gamma + \gamma\mathcal{U}_i) \mathcal{N}(\vect{e}_i, \sigma\vect{I})  + \gamma \sum_{v \neq i \in \mathcal{V}} \mathcal{U}_v \mathcal{N}(\vect{e}_v, \sigma\vect{I}),
\end{align}
where $\mathcal{U}_v$ is the unigram probability of word $v$ and $\sigma$ is small.
In other words, with probability $\gamma \mathcal{U}_v$, we replace the embedding $\vect{e}_i$ with 
another embedding sampled from a normal distribution centered at $\vect{e}_v$.

Note that noising the input word is equivalent to choosing 
a different word embedding vector to be used in a 
standard recurrent neural network.
Under the variational framework, we sample a 
different word embedding matrix for every sequence $t$ at training time, since
the integral is 
approximated with Monte Carlo by sampling from $q(\vect{e}_i)$ 
(Eq.~\ref{eq:sampling}).

The above formulation is related to word embedding dropout \citep{daile,iyyer,kumar}, although in word embedding dropout 
$q(\vect{e}_i) = (1 - \gamma) \mathcal{N}(\vect{e}_i, \sigma\vect{I})  + \gamma \mathcal{N}(\vect{0}, \sigma\vect{I})$.

\paragraph{$\mathbb{KL}$ term.}
For $p(\vect{e}_i) = \mathcal{N}(\vect{0}, \vect{I})$, we use 
Proposition 1 in \citet{klapproxdropout} and 
approximate the $\mathbb{KL}$ divergence between
a mixture of Gaussians $q(\vect{e}_i)$ and $p(\vect{e}_i)$ as:
\begin{align*}
\mathbb{KL}(q(\vect{e}_i) \Vert p(\vect{e}_i)) &\approx
\lambda \sum_{v \in \mathcal{V}} \frac{\theta_v^i}{2} \left(\vect{e}_v^{\top}\vect{e}_v + \mathbf{tr}(\sigma\vect{I}) - V(1 + \log 2\pi) -\log|\sigma\vect{I}| \right) \\
&= \lambda \sum_{v \in \mathcal{V}} \frac{\theta_v^i}{2} \Vert \vect{e}_{v} \Vert^2_2 + \text{constant},
\end{align*}
where $V$ is the vocabulary size and $\theta_v^i$ is the mixture proportion for word $i$: $(1 - \gamma + \gamma\mathcal{U}_i)$ for $v = i$ and $\gamma$ otherwise.
In practice, the $\mathbb{KL}$ term directly translates to an $\ell_2$ regularizer on 
each word embedding vector, but the regularization coefficient is data dependent.
More specifically, the regularization coefficient for word vector $\vect{e}_i$ taking into
account contributions from $\mathbb{KL}(q(\vect{e}_v) \Vert p(\vect{e}_v))$ for $v \in \mathcal{V}$ is:
\begin{align}
\label{eq:kl}
\lambda \frac{(V-1)\gamma + (1 - \gamma + \gamma\mathcal{U}_i)}{2} = \lambda \frac{1 - (V-2+\mathcal{U}_i)\gamma}{2}
\end{align}
In other words, the variational formulation of data smoothing results in a regularization coefficient that is a function of corpus statistics (the unigram distribution).

\paragraph{Other parameters.}
For other parameters $\vect{R}$ and $\vect{O}$, we can 
use either simple variational distributions such as
$q(\vect{R}) \sim \mathcal{N}(\vect{R}, \sigma \vect{I})$ and
$q(\vect{O}) \sim \mathcal{N}(\vect{O}, \sigma \vect{I})$, which become standard $\ell_2$ regularizers on these parameters;
or incorporate dropout by setting $q(\vect{r}_i) \sim (1 - \alpha) \mathcal{N}(\vect{r}_i, \sigma\vect{I})  + \alpha \mathcal{N}(\vect{0}, \sigma\vect{I})$, where $\alpha$ is the dropout probability and $\vect{r}_i$ is the $i$-th row of $\vect{R}$ \citep{vardropout}.

\paragraph{Training.} 
In the original noising formulation \citep{xie}, 
given a sequence $\textit{\{a, b, a, c\}}$, 
it is possible to get a noised sequence $\textit{\{\underline{d}, b, a, c\}}$, 
since the decision to noise at every timestep is independent of others.
In the variational framework, 
we use Monte Carlo integration 
$\vect{\tilde{W}} \sim q(\vect{W})$ for 
each sequence to compute 
$p(x_t \mid x_{<t}; \vect{\tilde{W}})$.
We can either sample one embedding matrix $\vect{E}$
per sequence similar to \citet{vardropout},
or sample one embedding matrix per timestep \citep{pushing}.
While the first method is computationally more efficient (we use it in our experiments),
if we decide to noise the first $a$ to $d$, 
we will have a noised sequence where every $a$ is replaced by $d$: $\textit{\{\underline{d}, b, \underline{d}, c\}}$.
At training time, we go through sequence $x_{<t}$ multiple times (once per epoch)
to get different noised sequences.

\paragraph{Predictions.}
For predictions, \citet{xie} do not noise any input.
In the variational framework, this corresponds to taking the mode of the variational distribution,
since $1-\gamma$ is almost always greater than $\gamma$.
They also reported that they did not observe any improvements by sampling.
Using the mode is rather uncommon in Bayesian RNNs.
A standard approach is to take the mean, so we can set:
\begin{align*}
\vect{\bar{e}}_i = (1 - \gamma + \gamma\mathcal{U}_i)\vect{e}_i + \sum_{v \neq i \in \mathcal{V}} \gamma\mathcal{U}_v \vect{e}_v.
\end{align*}

\paragraph{Extension to absolute discounting.}
It is straightforward to extend linear interpolation smoothing to absolute discounting by setting $\gamma_i = \gamma\frac{\text{distinct}(i, \bullet)}{\text{count}(i)}$ in Eq.~\ref{eq:vardist} and Eq.~\ref{eq:kl} above.

\subsection{Kneser-Ney Smoothing}
\label{sec:knsmooth}
We now consider the variational analog of Kneser-Ney data noising.
Instead of smoothing towards lower order $n$-grams, 
Kneser-Ney uses models that take into account contextual diversity.
For example, for bigram Kneser-Ney smoothing, 
we replace the unigram distribution with
$\mathcal{K}_i = \frac{\text{distinct}(\bullet, i)}{\sum_{v \in \mathcal{V}} \text{distinct}(\bullet, v)}$, where $\text{distinct}(\bullet, i)$ denotes the number of distinct bigrams that end with word $i$.
As a result, even if bigrams such as ``San Francisco'' and ``Los Angeles'' appear frequently in the corpus; ``Francisco'' and ``Angeles'' will not have high probabilities in $\mathcal{K}$ since they often follow ``San'' and ``Los''.

Recall that similar to absolute discounting, Kneser-Ney also
uses $\gamma_i = \gamma\frac{\text{distinct}(i, \bullet)}{\text{count}(i)}$.
However, if we replace an input word $x_{t-1}$, \citet{xie} proposed to also replace the corresponding output word $x_{t}$.
The intuition behind this is that since the probability of 
replacing an input word $x_{t-1}$ is proportional to the number of distinct bigrams that start with $x_{t-1}$, when we replace the input word, 
we also need to replace the output word (e.g., if we replace ``San'' by a word sampled from $\mathcal{K}$, 
we should also replace ``Francisco'').

\paragraph{Word embedding matrix $\vect{E}$.}
To get (bigram) Kneser-Ney smoothing, we use the following mixture of Gaussians variational distribution for $\vect{e}_i$:
\begin{align}
\label{eq:invar}
\vect{e}_i \sim q(\vect{e}_i) &= (1 - \gamma_i) \mathcal{N}(\vect{e}_i, \sigma\vect{I})  + \gamma_i \sum_{v \in \mathcal{V}} \mathcal{K}_v \mathcal{N}(\vect{e}_v, \sigma\vect{I}) \nonumber \\
&= (1 - \gamma_i + \gamma_i\mathcal{K}_i) \mathcal{N}(\vect{e}_i, \sigma\vect{I})  + \gamma_i \sum_{v \neq i \in \mathcal{V}} \mathcal{K}_v \mathcal{N}(\vect{e}_v, \sigma\vect{I}).
\end{align}

\paragraph{Output embedding matrix $\vect{O}$.}
For the output embedding matrix, we also use the same variational distribution as the word embedding matrix:
\begin{align}
\label{eq:outvar}
\vect{o}_i \sim q(\vect{o}_i) &= (1 - \gamma_i) \mathcal{N}(\vect{o}_i, \sigma\vect{I})  + \gamma_i \sum_{v \in \mathcal{V}} \mathcal{K}_v \mathcal{N}(\vect{o}_v, \sigma\vect{I}) \nonumber \\
&= (1 - \gamma_i + \gamma_i\mathcal{K}_i) \mathcal{N}(\vect{o}_i, \sigma\vect{I})  + \gamma_i \sum_{v \neq i \in \mathcal{V}} \mathcal{K}_v \mathcal{N}(\vect{o}_v, \sigma\vect{I}).
\end{align}

\paragraph{$\mathbb{KL}$ term.} 
Following similar derivations in the previous subsection,
it is straightforward to show that the approximated $\mathbb{KL}$ term introduces
the following regularization term to the overall objective:
\begin{align*}
\lambda \sum_{v \in \mathcal{V}} \frac{1 - (V - 2 + \mathcal{K}_v)\gamma_v}{2} \Vert \vect{e}_{v} \Vert_2^2 + \text{constant},
\end{align*}
and similarly for the output embedding matrix $\vect{O}$.

\paragraph{Training.} 
Recall that we use Monte Carlo integration 
$\vect{\tilde{W}} \sim q(\vect{W})$ for 
each sequence to compute 
$p(x_t \mid x_{<t}; \vect{\tilde{W}})$ at training time.
Since we need to noise the output word when the input word 
is replaced, we sample $\tilde{\gamma}_i$ at ${t-1}$ (for $x_{t-1} = i$)
and decide whether we will keep the original (input and output) words or replace them. 
If we decide to replace, we need to sample two new words 
from $\mathcal{K}_v \mathcal{N}(\vect{e}_v, \sigma\vect{I})$ and 
$\mathcal{K}_v \mathcal{N}(\vect{o}_v, \sigma\vect{I})$ respectively.
For other words $k$ that are not the target at the current timestep (i.e., $k \neq x_t$),
we can either assume that $\vect{\tilde{o}}_k = \vect{o}_k$ or alternatively sample 
(which can be expensive since we need to sample additional 
$V-1$ times per timestep).

\paragraph{Predictions}
For predictions, \citet{xie} also take 
the mode of the variational distribution (assuming $1-\gamma_i$ is almost always greater than $\gamma_i$).
We use the mean instead, similar to what we do in variational linear interpolation smoothing.

\subsection{Extensions}
\label{sec:variants}

Our derivations above provide insights into new methods
under the variational smoothing framework.
We describe three variants in the followings:

\paragraph{Tying input and output embeddings.}
\citet{tying} and \citet{press2017using} showed that tying the input and output embedding matrices (i.e., 
using the same embedding matrix for both) improves language modeling performance 
and reduces the number of parameters significantly.
We take inspirations from this work and sample both $\vect{O}$ and
$\vect{E}$ from the same base matrix.
As a result, we have fewer parameters to train due to this
sharing mechanism, but we still have different samples
of input and output embedding matrices per sequence.\footnote{Alternatively,
we could sample one matrix for both the input and output embeddings.
We found that this approach is slightly worse in our preliminary experiments.
}
Similar to previous results in language modeling \citep{tying, melisgl}, 
our experiments demonstrate 
that tying improves the performance considerably.

\paragraph{Combining smoothing and dropout.}
We can combine variational smoothing and variational dropout by modifying the variational distribution to incorporate a standard Gaussian component:
\begin{align*}
\vect{e}_i \sim q(\vect{e}_i) &= (1-\alpha) \left\{(1 - \gamma_i) \mathcal{N}(\vect{e}_i, \sigma\vect{I})  + \gamma_i \sum_{v \in \mathcal{V}} \mathcal{K}_v \mathcal{N}(\vect{e}_v, \sigma\vect{I})\right\} + \alpha 
\mathcal{N}(0, \sigma\vect{I}),
\nonumber
\end{align*}
where $\alpha$ is the dropout probability. Note that in this formulation, we either drop
an entire word by setting its embedding vector to zero (i.e., similar to word embedding dropout and blank noising) 
or choose an embedding vector from the set of words in the vocabulary.

However, it is more common to apply dropout to each dimension of the input embedding and output embedding matrix.
In this formulation, we have:
\begin{align}
\label{eq:element}
e_{i,j} \sim q(e_{i,j}) &= (1-\alpha_{i,j}) \left\{(1 - \gamma_i) \mathcal{N}(e_{i,j}, \sigma)  + \gamma_i \sum_{v \in \mathcal{V}} \mathcal{K}_v \mathcal{N}(e_{v,j}, \sigma)\right\} + \alpha_{i,j}
\mathcal{N}(0, \sigma).
\end{align}
At training time, while we sample $\tilde{\alpha}_{i,j}$ 
multiple times (once per word embedding dimension for each word),
we only sample $\tilde{\gamma}_i$ once per word to ensure that when the 
element is not noised, we still use the same base embedding vector.
We use this variant of smoothing and dropout throughout our experiments for our models.

\paragraph{Element-wise smoothing.}
The variational formulation above allows us to
derive a variant of data smoothing that samples each element 
of the embedding vector $\vect{e}_i$ independently (and similarly for $\vect{o}_i$).
Consider the variational distribution in Eq.~\ref{eq:element}.
At training time, if we sample both $\tilde{\alpha}_{i,j}$ 
and $\tilde{\gamma}_{i,j}$ multiple times (once per word embedding dimension for each word)
we arrive at a new element-wise smoothing method.
The main difference between this model and the previous model is that
each dimension in the input (and output) embedding vector is sampled independently.
As a result, the vector that is used is 
a combination of elements from various word vectors.
Notice that the mean under this new scheme is still the same as sampling per vector, so we do not need to change anything at test time.
One major drawback about this model is that it is computationally expensive since we need
to sample each element of each embedding vector.

\section{Experiments}
\label{sec:exp}
\subsection{Setup}
We evaluate our approaches on two standard language modeling datasets: Penn Treebank (PTB) and Wikitext-2.
We use a two-layer LSTM as our base language model.
We perform non-episodic training with batch size 64 using RMSprop \citep{rmsprop} as our optimization method.
We tune the RMSprop learning rate and $\ell_2$ 
regularization hyperparameter $\lambda$ for all models 
on a development set by a grid search 
on $\{0.002, 0.003, 0.004\}$ and $\{10^{-4},10^{-3}\}$ respectively, and use
perplexity on the development set to choose the best model.
We also tune $\gamma$ from $\{0.1, 0.2, 0.3, 0.4\}$.
We use recurrent dropout \citep{recdropout} for $\vect{R}$ and set it to 0.2,
and apply (element-wise) input and output embedding dropouts for $\vect{E}$ and $\vect{O}$ and set it to 0.5 when $\vect{E}, \vect{O} \in \mathbb{R}^{V\times512}$ and 0.7 when $\vect{E}, \vect{O} \in \mathbb{R}^{V\times1024}$ based on preliminary experiments.
We tie the input and output embedding matrices in all our experiments 
(i.e., $\vect{E} = \vect{O}$), except for the vanilla 
LSTM model, where we report results for both tied and untied.\footnote{Our preliminary experiments are consistent with previous work \citep{tying,melisgl,smerity} that show tying the input and output embedding matrices results in better models with fewer numbers of parameters.}

\subsection{Models}
We compare the following methods in our experiments:
\begin{itemize}
\item \textbf{Baseline}: a vanilla LSTM language model. We evaluate two variants: with tied input and output embeddings and without.
\item \textbf{Data noising (\textsc{DN})}: an LSTM language model trained with data noising using linear interpolation smoothing or bigram Kneser-Ney smoothing \citep{xie}.
\item \textbf{Variational smoothing (\textsc{VS})}: an LSTM language model with variational smoothing using linear interpolation or Kneser-Ney. For both models, we use the mean of the variational distribution at test time.\footnote{We discuss results
using sampling at test time in \S{\ref{sec:discussion}}.}
\item \textbf{Variational element-wise smoothing}: for the smaller PTB dataset, we evaluate an LSTM language model that uses elementwise Kneser-Ney variational smoothing and dropout. We also use the mean at test time.
\end{itemize}

\renewcommand{\arraystretch}{1.2}
\begin{table}[t]
\small
\vspace{-0.4cm}
    \centering
    \begin{tabular}{|l|c|c|r|r|r|r|}
     \hline
      \multirow{2}{*}{\textbf{Model}} & \textbf{LSTM} & \# of & \multicolumn{2}{c|}{\textbf{PTB}} & \multicolumn{2}{c|}{\textbf{Wikitext-2}} \\
      \cline{4-7}
       & \textbf{hidden size} & \textbf{params.} & \textbf{Dev} &  \textbf{Test} & \textbf{Dev} & \textbf{Test} \\
       \hline
       \hline
       Vanilla LSTM \citep{xie}  & 512 & - & 84.3 & 80.4 & - & -\\
       Vanilla LSTM \citep{xie} & 1500 & - & 81.6 & 77.5 & - & - \\
       \textsc{DN}: Kneser-Ney \citep{xie} & 512 & - & 79.9 & 76.9 & - & -\\
       \textsc{DN}: Kneser-Ney \citep{xie} & 1500 & - & 76.2 & 73.4 & - & -\\
       Var. dropout \citep{vardropout} & 1500 & - & - & 73.4 & - & - \\
       \hline
       \hline
       Vanilla LSTM: untied & 512 & 14M/38M & 89.6 & 84.5 & 106.3 & 100.8 \\
       \hline
       Vanilla LSTM: tied & \multirow{3}{*}{512} & \multirow{3}{*}{9M/21M} & 80.0 & 74.0 & 90.6 & 86.6 \\
       \textsc{DN}: linear interpolation & & & 79.4 & 73.3 & 88.9 & 84.6\\
       \textsc{DN}: Kneser-Ney & & & 75.0 & 70.7 & 86.1 & 82.1 \\
       \hline
       \textsc{VS}: linear interpolation & \multirow{4}{*}{512}& \multirow{4}{*}{9M/21M}& 76.3 & 71.2 & 84.0 & 79.6\\
       \textsc{VS}: Kneser-Ney & & & 74.5 & 70.6 & 84.9 & 80.9 \\
       \textsc{VS}: element-wise & & &70.5 &66.8 & -&-\\
       \hline
       \hline
       Vanilla LSTM: untied & 1024 & 37M/85M & 90.3 & 85.5 & 97.6 & 91.9 \\
       \hline
       Vanilla LSTM: tied & \multirow{3}{*}{1024} & \multirow{3}{*}{27M/50M} & 75.9 & 70.2 & 85.2 & 81.0\\
       \textsc{DN}: linear interpolation & & & 75.5 & 70.2 & 84.3 & 80.1\\
       \textsc{DN}: Kneser-Ney & & & 71.4 & 67.3 & 81.9 & 78.3\\
       \hline
       \textsc{VS}: linear interpolation & \multirow{4}{*}{1024}& \multirow{4}{*}{27M/50M} & 71.7 & 67.8 & 80.5 & 76.6\\
       \textsc{VS}: Kneser-Ney & &&70.8 & 66.9 & 80.9 & 76.7\\
       \textsc{VS}: element-wise & & & 68.6 & 64.5 & -&-\\
       \hline
    \end{tabular}
    \caption{Perplexity on PTB and Wikitext-2 datasets. \textsc{DN} and \textsc{VS} 
    denote data noising and variational smoothing. The two numbers (*M/*M) in the 
    \# of params. column denote the number of parameters for PTB and Wikitext-2 respectively.}
    \label{tbl:results}
\end{table}

\vspace{-0.1cm}
\subsection{Results}
Our results are summarized in Table~\ref{tbl:results}.
Consistent with previous work on 
tying the input and output embedding matrices in language models 
\citep{tying,melisgl,smerity}, 
we see a large reduction in perplexity (lower is better)
when doing so.
While our numbers are generally better,
the results are also consistent with \citet{xie} that show 
linear interpolation data noising is slightly better than vanilla LSTM with dropout,
and that Kneser-Ney data noising outperforms these methods
for both the medium (512) and large (1024) models.

Variational smoothing improves over data noising in all cases, both for
linear interpolation and Kneser-Ney.
Recall that the main differences between variational smoothing and data noising are:
(1) using the mean at test time, (2) having a data dependent $\ell_2$ regularization coefficient
that comes from the $\mathbb{KL}$ term,\footnote{The data dependent $\ell_2$
coefficient penalizes vectors that are sampled more often higher.}
and (3) how each method interacts with the input and output embedding tying mechanism.\footnote{
In the variational framework, if we 
sample one matrix for the input and output embeddings, 
it effectively noises the output words even for linear interpolation.
If we sample two matrices from the same base matrix,
these matrices can be different at training time even if the parameters are tied. 
As described in \S{\ref{sec:variants}}, we
use the latter in our experiments.
}
Our results suggest that the 
choice of the proposal distribution to sample 
from is less important for variational smoothing.
In our experiments,
Kneser-Ney outperforms linear interpolation on PTB 
but linear interpolation is slightly better on Wikitext-2.

Element-wise variational smoothing performs the best for both small 
and large LSTM models on PTB.
We note that this improvement comes at a cost, since this method is computationally expensive.\footnote{Element-wise dropout can be implemented efficiently by sampling a mask (zero or one with some probability) and multiply the entire embedding matrix with this mask. In element-wise smoothing, we need to sample an index for each dimension and reconstruct the embedding matrix for each timestep.}
It took about one day to train the smaller model as opposed to a couple hours
without element-wise smoothing.
As a result, we were unable to train this on a much bigger dataset such as Wikitext-2.
Nonetheless, our results show that applying smoothing in the embedding
(latent) space results in better models.

\subsection{Discussions}
\label{sec:discussion}
\paragraph{Sampling at test time.} In order to better understand another prediction method for these models,
we perform experiments where we sample at test time for both data noising and variational smoothing
(instead of taking the mode or the mean).
We use twenty samples and average the log likelihood.
Our results agree with \citet{xie} that mentioned sampling does not provide
additional benefits.
In our experiments, the perplexities 
of 512- and 1024-dimensional \textsc{DN}-Kneser-Ney models
increase to 96.5 and 85.1 (from 75.0 and 71.4)
on the PTB validation set.
For \textsc{VS}-Kneser-Ney models, the perplexities increase 
to 89.8 and 78.7 (from 74.5 and 70.8).
Both our results and \citet{xie} suggest that 
introducing data dependent noise at test time 
is detrimental for recurrent neural network language models.

\begin{figure}[t]
\centering
\includegraphics[scale=0.35]{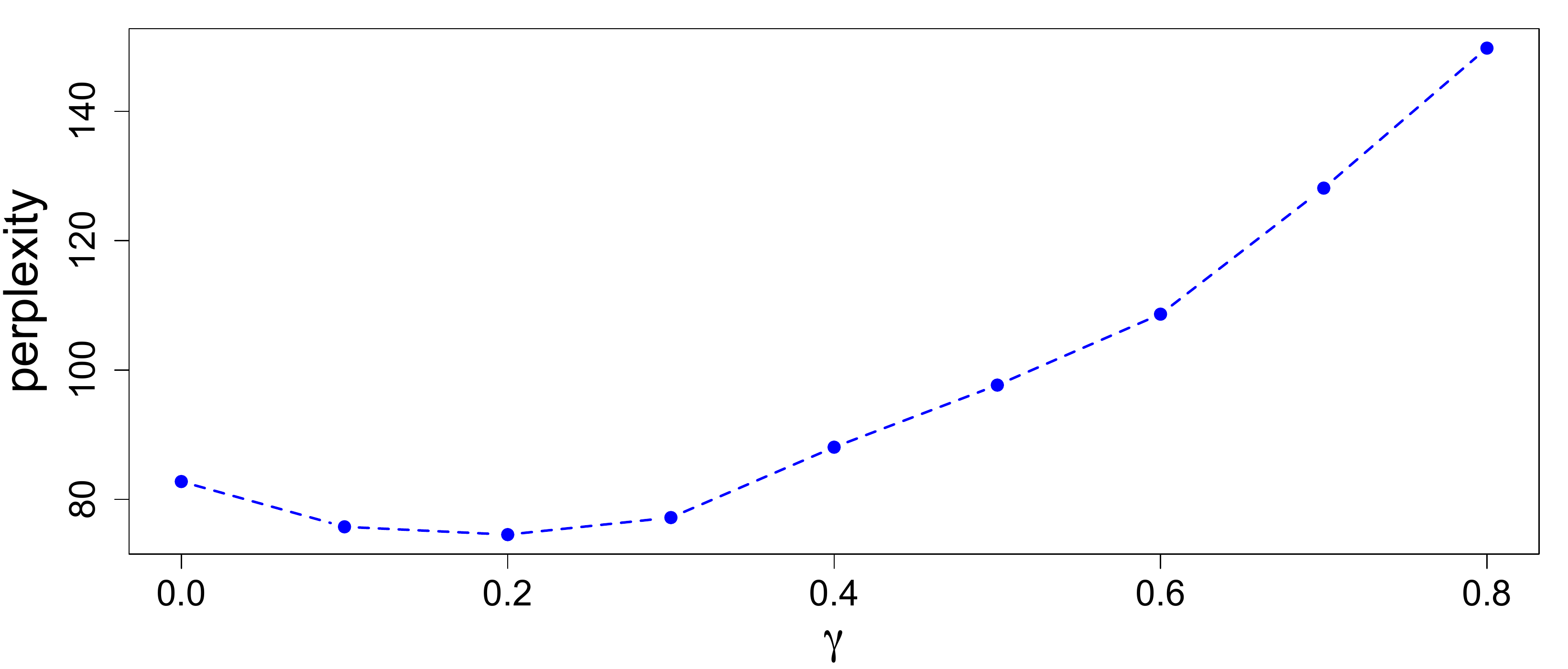}
\caption{Validation set perplexities on PTB for \textsc{VS}: Kneser-Ney (512 dimensions).}
\label{fig:gamma}
\end{figure}

\paragraph{Sensitivity to $\gamma$.} We evaluate the sensitivity of variational smoothing 
to hyperparameter $\gamma$.  
Figure~\ref{fig:gamma} shows perplexities on the PTB validation set for 
a variant of our models.
The model generally performs well within the range of 0.1 and 0.3,
but it becomes progressively worse as we increase $\gamma$
since there is too much noise.


\paragraph{Other applications.} While we focus on language modeling in this paper, the proposed technique is
applicable to other language processing tasks.
For example, \citet{xie} showed that data smoothing improves machine translation, so our techniques can be used in that setup as well. 
It is also interesting to consider how variational smoothing interacts with and/or can be applied to memory augmented language models \citep{tran,grave,yogatama} and state-of-the-art language models \citep{yang2017breaking}.
We leave these for future work.


\section{Conclusion}
We showed that data noising in recurrent neural network
language models can be understood as a Bayesian recurrent
neural network with a variational distribution that consists of a mixture of Gaussians whose mixture weights are a function of the proposal distribution used in data noising (e.g., the unigram distribution, the Kneser-Ney continuation distribution).
We proposed using the mean of the variational distribution at prediction time as a better alternative to using the mode.
We combined it with variational dropout, presented two extensions (i.e., variational smoothing with
tied input and output embedding matrices and element-wise smoothing), and demonstrated
language modeling improvements on Penn Treebank and Wikitext-2.

\bibliography{iclr2019_conference}
\bibliographystyle{iclr2019_conference}

\end{document}